\documentclass[pdflatex,sn-mathphys-num]{sn-jnl}

\usepackage{graphicx}%
\usepackage{multirow}%
\usepackage{amsmath,amssymb,amsfonts}%
\usepackage{amsthm}%
\usepackage{mathrsfs}%
\usepackage[title]{appendix}%
\usepackage{xcolor}%
\usepackage{textcomp}%
\usepackage{manyfoot}%
\usepackage{booktabs}%
\usepackage{algorithm}%
\usepackage{algorithmicx}%
\usepackage{algpseudocode}%
\usepackage{listings}%
\usepackage{booktabs}
\usepackage{tabularx}

\theoremstyle{thmstyleone}%
\theoremstyle{thmstyletwo}%

\theoremstyle{thmstylethree}%

\begin{document}

\title{Susceptible Reservoir Architectures for Regime-Conditional Volatility Forecasting}


\author*[1]{\fnm{Aliaksei} \sur{Kaliutau}}
\email{\href{mailto:aliaksei.kaliutau24@alumni.imperial.ac.uk}{aliaksei.kaliutau24@alumni.imperial.ac.uk}}

\affil[1]{\orgname{Monodromy}, \orgaddress{\city{London}, \country{United Kingdom}}}


\abstract{
Volatility forecasting is dominated by persistence and measurement noise, leaving limited residual structure for nonlinear models to exploit. We introduce Susceptible Architectures (SUSA), a reservoir-design principle for volatility forecasting, and its two concrete implementations, based on complex-valued open-chain and periodic reservoirs and regime-conditioned experts to interpret reservoir features across calm, onset, recovery, and persistent-stress states. We also implement open-system $q$-qubit counterparts in Qiskit while retaining a common AR-Ridge anchor and a bounded residual correction trained under QLIKE.

We evaluate models on 16 U.S. equity and exchange-traded-fund series using three disjoint chronological training, validation, and test folds, a 12-observation input window, and a five-observation forecast horizon.

The proposed models perform competitively with GARCH, achieving statistically significant QLIKE improvements for specific assets (IWM, XLP). Also models' forecasts complement HARQ-style predictions: a stacked ensemble improves mean QLIKE by 0.0116 over its strongest constituent and wins in 75\% of test scenarios.}




\maketitle

\section{Introduction}\label{sec1}

Volatility forecasting is an unusually hard task in for ML models. The target is often persistent, heteroskedastic, heavy-tailed, and regime dependent. 

There are several strong classical models that explain a large part of the log volatility with acceptable accuracy. For example, GARCH encodes conditional variance recursion~\cite{Bollerslev1}; HAR represents heterogeneous daily, weekly, and monthly persistence~\cite{Corsi1}; HARQ-style version of HAR focusing on the reliability of realized-variance measurements~\cite{Bollerslev2}. Any competitive non-linear model should beat all these baselines and improve on the residual tail. However, due to the complexity of the target signal and the extremely low SNR ratio, this became a long-standing \emph{quaestio vexata} of the financial forecast.

In this work, we explore the deep connection between the susceptibility of complex dynamic systems and the market signals.

Susceptibility measures how strongly a system’s observable response changes when a parameter or structure is perturbed. Therefore, specifically designed critical systems can play the role of an amplifier. For example, near quantum critical points and driven-dissipative transitions, susceptibilities can become strongly enhanced because a small perturbation reorganizes a large part of the state. This is why critical systems are studied as resources for quantum sensing and metrology~\cite{DiCandia2023}. 

We want to use the susceptibility in computational sense rather than ontological: we do not assume that a financial market is a quantum system or that market regimes map directly to quantum states. We instead explore the idea that deliberately susceptible reservoirs can transform weak differences between volatility paths into structured contrasts that are easier for a low-capacity (usually a 1-layer) readout to detect.

The susceptibility of complex systems is a very generic and unifying concept. To narrow the scope and to define a starting point, we begin with quantum-inspired classical model, whose architecture can be easily re-formulated in quantum terms later. 

A generic reservoir is usually selected for memory, nonlinearity, expressivity, or state-space dimension. In comparison, a susceptible reservoir is selected for the geometry/dynamics of its response under matched interventions. In other words, we want to see and exploit the difference between the trajectories generated from the same input when we apply controlled changes to the reservoir. We call this design principle SUSceptible Architectures (SUSA).

SUSA generalizes a set of increasingly broad ideas. The narrowest form is parameter susceptibility: the reservoir is evaluated at nearby operating points, and their finite difference can be defined as a sensitivity along one physical parameter. The next form is structural susceptibility when a boundary pattern is changed. The broadest form is regime-conditional susceptibility: the same response contrast is interpreted differently according to the inferred state of the forecasting process.

In our initial experiments we studied reservoir tuned near quantum EP points and we found that parameter susceptibility has too many limitations to beat the strong baselines. In this paper, we focus on the boundary and regime susceptibility.

We start from the classically tractable complex reservoir that acts as a proxy, so we can quickly test the architectural contribution of directional transport, open-versus-periodic counterfactuals, and regime-conditioned readouts without invoking quantum dynamics. We then lift the same architecture to an open quantum reservoir. In the quantum version, we replace the $q$-node complex state with a $q$-qubit density operator evolving under coherent interactions and CPTP channels~\cite{Jaeger01,Mart23}. The resulting tensor-product state space supports a much richer family of coherence and many-body correlation responses. 

\begin{figure}[h]
\centering 
\includegraphics[width=0.6\textwidth]{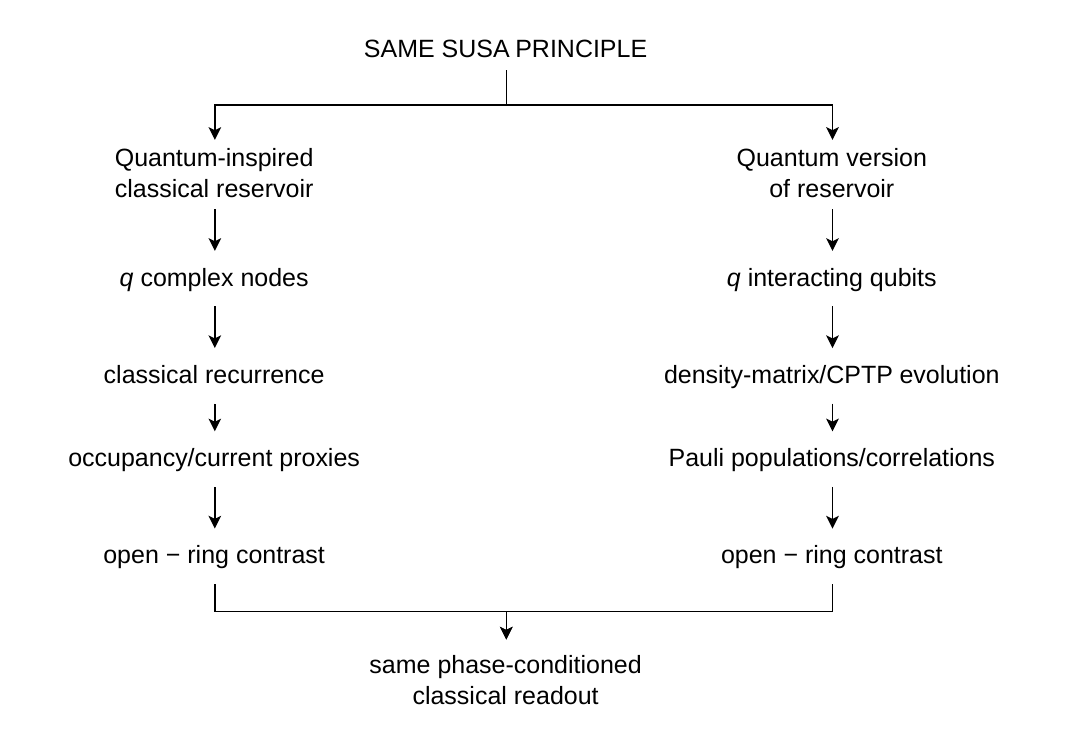}
\caption{We can design many different architectures, which are exploiting effectively the same idea}\label{fig:fig1}
\end{figure}

As we will show later, there is no single road to conduct this classical to quantum translation; instead we have to deal with potentially infinite number of variants. Enhanced susceptibility can occur in many places: near degeneracies, gap closings, exceptional points, bifurcations, or sharp changes in stationary state (Figure~\ref{fig:fig1}). 

In our design philosophy, the quantum version of QRC architecture is not a different forecasting idea - it is a richer physical version of the same susceptibility principle.
 
Note that the larger quantum state space only provides extra representational capacity, but it does not guaranty predictive advantage. Articles on expressivity analysis show that performance depends jointly on encoding, memory, size, and scrambling; larger reservoirs can become obstructive~\cite{Expressivity},~\cite{Li26}.

Recent work has applied quantum reservoir computing directly to realized-volatility forecasting~\cite{Li26}. Our contribution differs by constructing matched open-chain and periodic counterfactuals, interpreting their response contrast conditionally on market phase, and retaining a common classical persistence anchor.

In our paper, we explore and answer the following 2 research questions:

\begin{itemize}
    \item Can structural susceptibility extract stable residual information beyond econometric, autoregressive, and classical-reservoir baselines?
    \item Can phase-conditioned interpretation improve the accuracy and tail robustness of susceptibility features relative to a single global readout?
\end{itemize}

\section{Methods}\label{sec2}

\subsection{Core Architectures}

In this chapter, we design quantum-inspired models and their full quantum variants.

Boundary susceptibility is modeled under the open chain and close ring dynamics of the reservoir states. This idea is implemented in SC-RC, a quantum-inspired complex-valued reservoir that measures structural susceptibility through an open-versus-periodic counterfactual.

\begin{figure}[h]
\centering 
\includegraphics[width=0.7\textwidth]{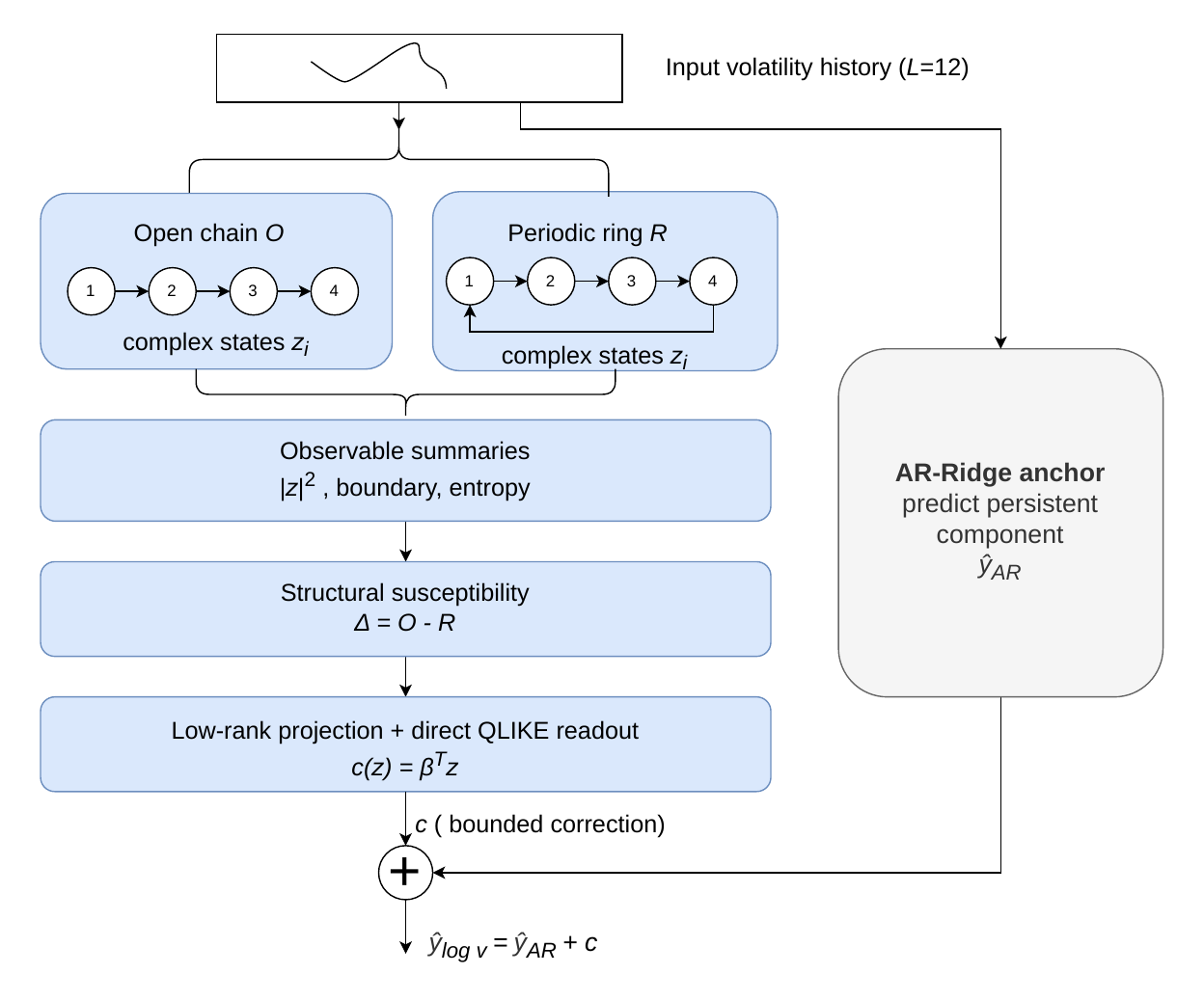}
\caption{The architecture of SC-RC. The input volatility window first enters a classical anchor branch and a susceptibility branch. The anchor branch produces the baseline log-variance forecast. The susceptibility branch divides into two matched reservoirs: an open chain and a periodic ring. In the quantum-inspired model, these blocks contain complex recurrent nodes; in the quantum model, they contain a density operator evolving under coherent interactions and CPTP channels. Both branches emit state and temporal observables. A contrast block computes their signed and normalized differences. A low-rank QLIKE readout converts this susceptibility representation into a bounded residual correction.}\label{fig:cflse}
\end{figure}

This reservoir contains two chains (each is a vector of complex node states). The input values modulate both the amplitude and phase of the injected signal. The open chain contains directional nearest-neighbor interactions with two physical boundaries. The periodic chain uses the same input masks, local dynamics, node count, and coupling construction but adds the closing edge that connects the last node to the first. The model then forms signed and normalized open-minus-ring contrasts. These features measure how strongly the encoded input path responds to the removal or insertion of a boundary connection (Figure~\ref{fig:cflse}).

\begin{figure}[htbp]
\centering 
\begin{minipage}{0.48\textwidth}
  \centering
  \includegraphics[width=\linewidth]{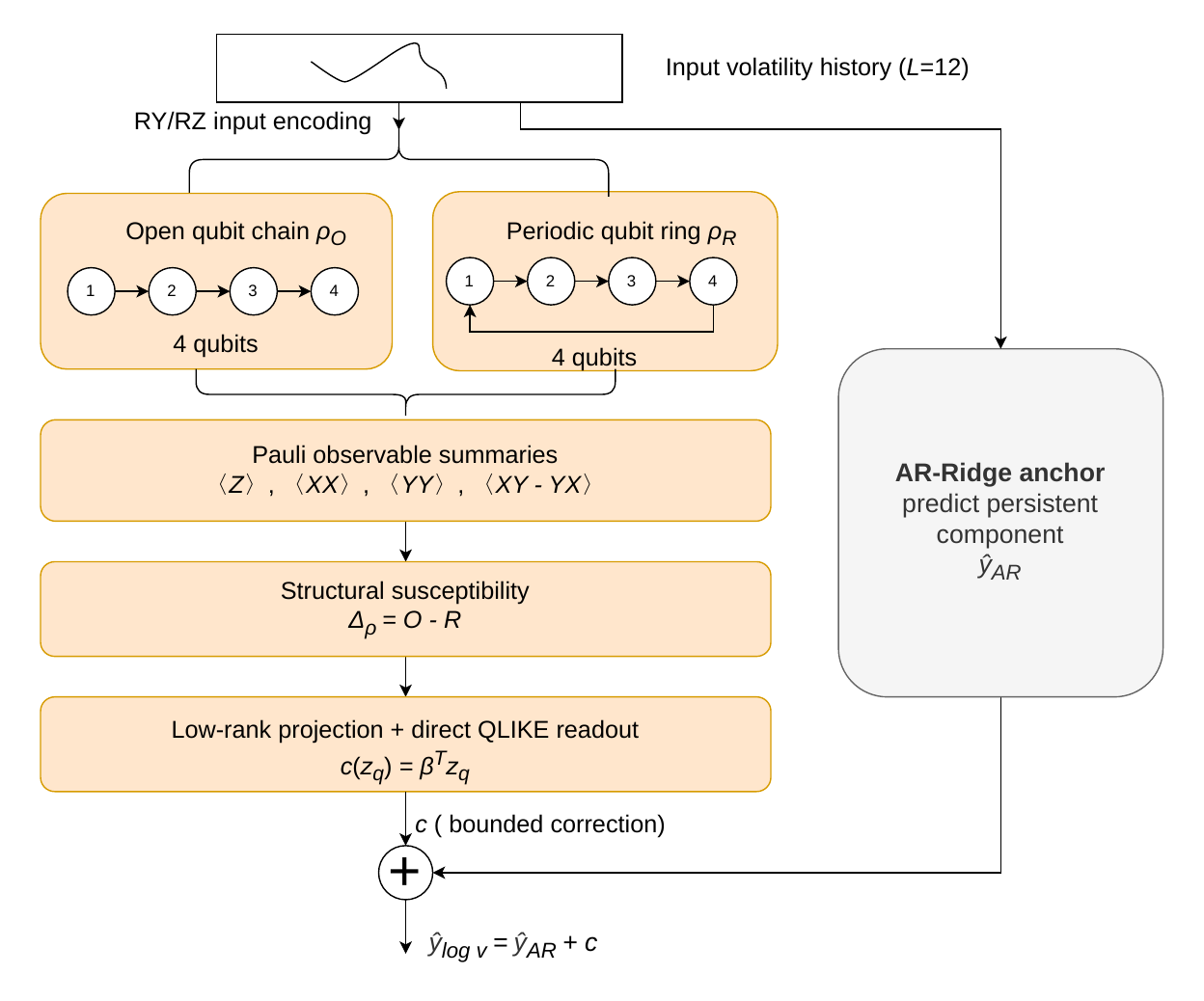}
  \caption{Architecture of SC-OSQRC.}\label{fig:osqrc}
\end{minipage}\hfill
\begin{minipage}{0.48\textwidth}
  \centering
  \includegraphics[width=\linewidth]{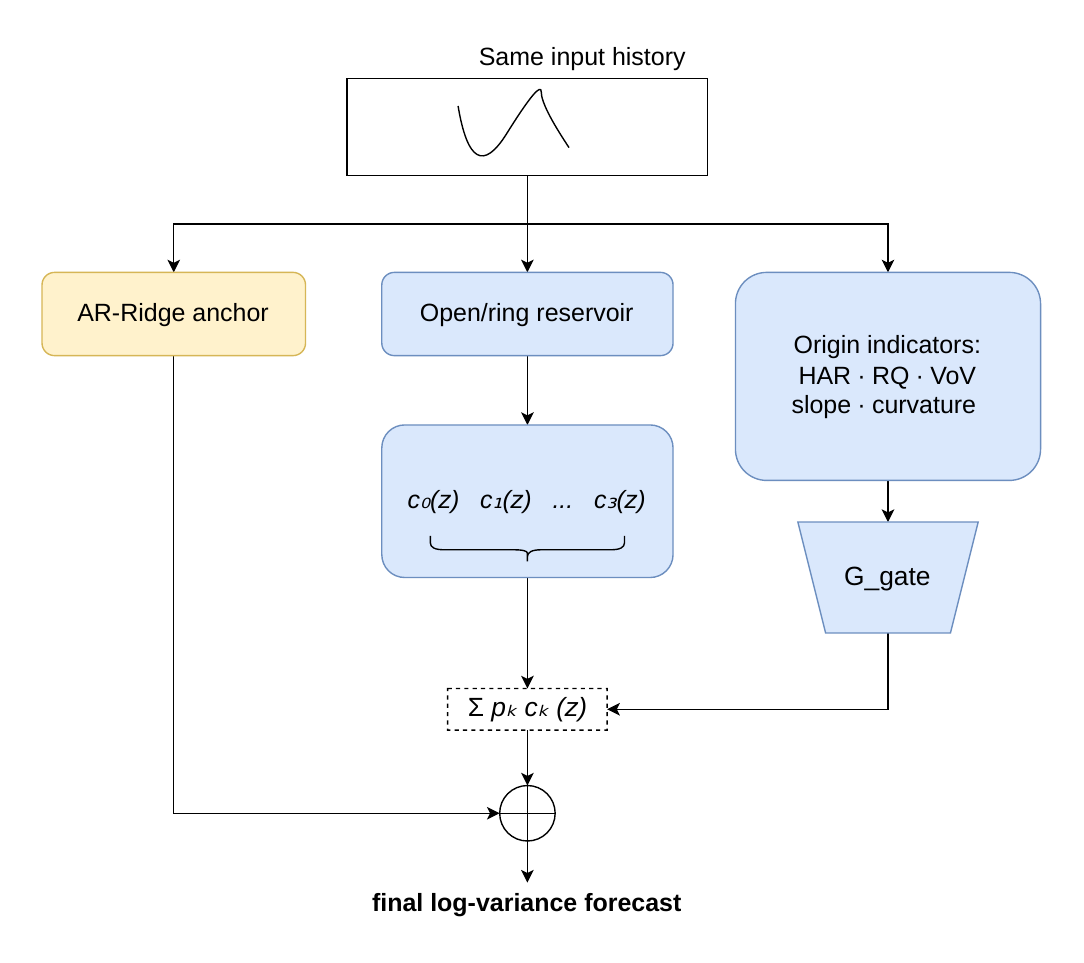}
  \caption{Architecture of PHASE-RC-MoE.}\label{fig:qrcmoe}
\end{minipage}
\end{figure}

In SC-OSQRC, the $q$-node complex state is replaced by the $q$-qubit density operator. Sequential inputs are encoded through local rotations. Coherent field terms and two-qubit interactions propagate information, while directional trace-preserving channels model asymmetric transfer and amplitude damping supplies fading memory (Figure~\ref{fig:osqrc}).

The quantum reservoir is observed through local Pauli populations, adjacent correlations, coherence proxies, directed-current observables, boundary imbalance, spatial concentration, and entropy-like summaries. Measurements collected across the input sequence produce final-state and temporal statistics. As in SC-RC, the model forms open-minus-ring and normalized contrast features, projects them to low rank, and learns a bounded residual correction trained by minimizing QLIKE over the same classical anchor.

PHASE-RC-MoE has a similar architecture, but it is complemented by applying a regime-conditioned readout. Several phase-specialized QLIKE readouts learn four transition states: calm, onset, recovery, and persistent stress (Figure~\ref{fig:qrcmoe}). The output is the probability-weighted combination of their outputs. Calm and persistent-stress experts can learn different mappings for otherwise similar reservoir responses, while onset and recovery experts can distinguish increasing from decreasing instability. On implementation level of PHASE-RC-MoE, we are using complex node states as well, so that this quantum-inspired architecture easily translatable into full quantum model. 

PHASE-OSQRC-MoE differs from PHASE-RC-MoE only in the presence of the quantum reservoir that supplies a fixed nonlinear temporal transformation, and optimization occurs in a small 4-head classical layer.

In both models, we do not try to predict the raw volatility from scratch. Instead, we use AR-Ridge as the common anchor and measure only small nonlinear incremental values beyond the raw volatility.

\subsection{Forecasting formulation}
\label{sec:formal_method}

For the input $s$ and forecast origin $t$, let
\begin{equation}
    x_{s,t-j}
    =
    \log\!\left(\max\{r_{s,t-j}^{2},\varepsilon\}\right),
    \qquad j=0,\ldots,L-1,
\end{equation}
and define the $H$-day future variance target
\begin{equation}
    v_{s,t}^{(H)}
    =
    \frac{1}{H}\sum_{h=1}^{H}r_{s,t+h}^{2},
    \qquad
    y_{s,t}=\log v_{s,t}^{(H)}.
    \label{eq:forecast_target}
\end{equation}
We use $L=12$ and $H=5$.

All proposed models preserve a common AR-Ridge persistence forecast and learn
only a bounded nonlinear correction:
\begin{equation}
    \widehat y_{s,t}
    =
    \widehat y^{\,\mathrm{AR}}_{s,t}
    +
    \operatorname{clip}
    \left(
        f_{\theta}(\mathbf z_{s,t}),-C,C
    \right),
    \qquad
    \widehat v_{s,t}=\exp(\widehat y_{s,t}).
    \label{eq:anchored_forecast}
\end{equation}
The readout is fitted directly under QLIKE,
\begin{equation}
    \widehat{\theta}
    =
    \arg\min_{\theta}
    \sum_{i\in\mathcal I_{\mathrm{OOF}}}
    \left[
        \widehat y^{\,\mathrm{AR,OOF}}_i
        +f_{\theta}(\mathbf z_i)
        +v_i
        \exp\!\left(
            -\widehat y^{\,\mathrm{AR,OOF}}_i
            -f_{\theta}(\mathbf z_i)
        \right)
    \right]
    +
    \lambda\|\theta\|_2^2,
    \label{eq:qlike_readout}
\end{equation}
where the anchor predictions used for readout fitting are generated
chronologically out of sample.

\subsection{Structural-susceptibility reservoir}
\label{sec:sc_rc}

The same encoded input path is propagated through two matched reservoirs:
an open chain and a periodic ring. The reservoirs share their node count,
input masks, random realization, coupling scale, and spectral radius; the ring
differs only by the wrap connection between the first and last nodes.

For topology $b\in\{o,r\}$, the classical complex state evolves as
\begin{equation}
    \mathbf s_{t,\ell}^{(b)}
    =
    \varphi\!\left(
        \mathbf W_b\mathbf s_{t,\ell-1}^{(b)}
        +
        \mathbf d(x_t)
    \right),
    \qquad
    \varphi(z)=\tanh(|z|)e^{\mathrm i\arg z}.
    \label{eq:classical_reservoir}
\end{equation}
Here $\mathbf W_o$ contains nearest-neighbour open-chain couplings, whereas
$\mathbf W_r$ additionally contains the periodic wrap edge.

Let $\mathbf o_{s,t}^{(o)}$ and $\mathbf o_{s,t}^{(r)}$ denote matched
summaries of the two trajectories. The proposed susceptibility representation
is
\begin{align}
    \boldsymbol\Delta_{s,t}
    &=
    \mathbf o_{s,t}^{(o)}
    -
    \mathbf o_{s,t}^{(r)},\\
    \widetilde{\boldsymbol\Delta}_{s,t}
    &=
    \frac{
        \boldsymbol\Delta_{s,t}
    }{
        |\mathbf o_{s,t}^{(o)}|
        +
        |\mathbf o_{s,t}^{(r)}|
        +\varepsilon_z
    },\\
    \mathbf z_{s,t}^{\mathrm{SC}}
    &=
    \left[
        \boldsymbol\Delta_{s,t};
        \widetilde{\boldsymbol\Delta}_{s,t}
    \right].
    \label{eq:sc_representation}
\end{align}
Thus, the learned correction depends on the response of one encoded path to a
controlled change in boundary topology rather than on the absolute state of a
single reservoir.

\subsection{Open-system quantum realization}
\label{sec:osqrc}

The quantum counterpart replaces the complex state by a density operator
$\rho_t^{(b)}$ over $q$ qubits. For each encoded observation,
\begin{equation}
    \rho_t^{(b)}
    =
    \mathcal A_{\gamma}
    \circ
    \mathcal T_{E_b}
    \circ
    \mathcal U_{E_b}
    \circ
    \mathcal U_{\mathrm{in}}(x_t)
    \left[
        \rho_{t-1}^{(b)}
    \right],
    \label{eq:quantum_reservoir}
\end{equation}
where $\mathcal U_{\mathrm{in}}$ performs data-dependent local rotations,
$\mathcal U_{E_b}$ applies coherent $XX+YY$ interactions,
$\mathcal T_{E_b}$ implements directional excitation transfer, and
$\mathcal A_{\gamma}$ supplies amplitude damping.

The topology edge sets are
\begin{equation}
    E_o=\{(1,2),\ldots,(q-1,q)\},
    \qquad
    E_r=E_o\cup\{(q,1)\}.
    \label{eq:quantum_topologies}
\end{equation}
The two systems otherwise use identical input encoding, interaction,
dissipation, initial state, and measurement coordinates.

Local populations and adjacent coherence/current observables are summarized
over the input window. The quantum susceptibility representation applies the
same signed and normalized open--ring contrasts as
Equation~\eqref{eq:sc_representation}. The quantum reservoir is a fixed
feature generator; the anchor, projection, QLIKE readout, and phase gate
remain classical.

\subsection{Phase-conditioned mixture of experts}
\label{sec:phase_moe}

We distinguish four volatility transitions using the current and future
stress indicators
\begin{equation}
    a_{s,t}
    =
    \mathbf{1[h_{s,t}\ge q_{s,k}]},
    \qquad
    b_{s,t}
    =
    \mathbf{ 1[y_{s,t}\ge q_{s,k}]},
    \label{eq:phase_indicators}
\end{equation}
where $q_{s,k}$ is estimated from the training segment of asset $s$ and fold
$k$. The pairs $(a,b)=(0,0),(0,1),(1,0),(1,1)$ define calm, onset,
recovery, and persistent-stress observations.

The future-dependent indicator $b_{s,t}$ is used only to construct training
labels. At forecast time, a causal gate produces phase probabilities
\begin{equation}
    \boldsymbol\pi_{s,t}
    =
    \operatorname{softmax}
    \left(
        \mathbf A\mathbf g_{s,t}+\mathbf a
    \right),
    \label{eq:phase_gate}
\end{equation}
from trailing volatility, quarticity, jump, drawdown, and volatility-of-volatility features $\mathbf g_{s,t}$.

The final correction is a probability-weighted combination of four
phase-specialized QLIKE heads:
\begin{equation}
    \widehat y_{s,t}^{\,\mathrm{PHASE}}
    =
    \widehat y_{s,t}^{\,\mathrm{AR}}
    +
    \operatorname{clip}
    \left(
        \sum_{m=0}^{3}
        \pi_{s,t,m}
        f_{\theta_m}(\mathbf z_{s,t}),
        -C,C
    \right).
    \label{eq:phase_forecast}
\end{equation}
PHASE-RC-MoE and PHASE-OSQRC-MoE differ only in whether
$\mathbf z_{s,t}$ is produced by the classical or open-system quantum
reservoir.

\subsection{Testing methodology}

We chose a set of canonical and structurally relevant benchmarks: AR-Ridge for linear persistence, GARCH(1,1) for conditional-variance recursion, HAR/HARQ-style models for heterogeneous
multi-horizon persistence. ESN is the closest conventional reservoir comparator because it also uses fixed recurrent dynamics followed by a trained linear readout.

Autoregressive Ridge (AR, regularized linear persistence baseline) is used as an anchor and a basic baseline. We use AR as an absolute baseline because it provides a transparent reference point; the relative performance of GARCH- and HARQ-style models can vary depending on market (Table~\ref{tab:table1}).

\begin{table}[htbp]
    \centering
    \begin{tabular}{l|p{9cm}}
        \hline
        Comparison & Scientific meaning \\
        \hline
        QRC vs AR-Ridge & Does the reservoir add nonlinear temporal information beyond lagged log variance? \\
        QRC vs HAR/HARQ & Does it add value beyond volatility-specific multi-horizon features and stress proxies? \\
        QRC vs GARCH & Does it add value beyond a standard conditional-variance recursion? \\
        \hline
    \end{tabular}
    \caption{The benchmark is designed as a hierarchy rather than a flat leaderboard.}
    \label{tab:table1}
\end{table}

For our experiments, we selected 16 symbols, representing different U.S. equity exposures (SPY, QQQ, IWM, XLB, XLE, XLF, XLI, XLK, XLP, XLU, XLV, XLY, AAPL, MSFT, JPM, NVDA).

For each symbol, we construct a versioned daily market dataset from canonicalized OHLCV observations, then calculate close-to-close log returns, squared returns as a daily realized-variance proxy, and causal trailing volatility features. We split the original training data into 3 folds, each of which is split into training/test windows ($L$=12,$H$=5). Thus, the input window for models is previous 12 log-variance observations, and the target is mean variance over the horizon of next 5 observations.

\begin{table}[t]
\centering
\label{tab:chronological_folds}
\scriptsize
\setlength{\tabcolsep}{3.4pt}
\renewcommand{\arraystretch}{1.08}
\begin{tabular}{@{}c cc cc cc@{}}
\toprule
& \multicolumn{2}{c}{Training}
& \multicolumn{2}{c}{Validation}
& \multicolumn{2}{c}{Test} \\
\cmidrule(lr){2-3}
\cmidrule(lr){4-5}
\cmidrule(l){6-7}
Fold
& First & Last
& First & Last
& First & Last \\
\midrule
1
& 2006-01-26 & 2010-02-16
& 2010-02-23 & 2011-02-24
& 2011-03-03 & 2012-11-08 \\
2
& 2012-11-15 & 2016-12-05
& 2016-12-12 & 2017-12-14
& 2017-12-21 & 2019-09-03 \\
3
& 2019-09-10 & 2023-09-28
& 2023-10-05 & 2024-10-09
& 2024-10-16 & 2026-06-30 \\
\bottomrule
\end{tabular}
\caption{Chronological outer folds used for model selection and evaluation.
Dates are inclusive. Purge intervals separate the training, validation,
and test segments. Folds have very different market conditions.}

\end{table}

Dataset quality and leakage are controlled at both dataset-construction level and evaluation stages. Variance smoothing and all features are computed using trailing information only, and no observation from the forecast horizon is included in its corresponding input window (i.e. we do not use centered mean calculation, etc). The standardization parameters and the stress threshold are estimated from training observations only and subsequently applied without refitting to validation or test data. Handcrafted HAR, quarticity, draw-down, and gating variables are aligned to the forecast origin rather than the target date, ensuring that every model input would have been observable when the forecast was issued. We validate strictly positive variance targets, monotonically ordered dates, contiguous chronological source splits, and stores file hashes.  

The dataset for each symbol is divided into disjoint chronological outer folds. Each fold contains its own training/validation/test segment, with purge intervals inserted between adjacent segments and between outer folds. 

We quantify uncertainty in loss differentials using dependence-aware confidence intervals and tests of equal predictive accuracy. We use the Holm-Bonferroni adjustment~\cite{holm1979} as a multi-comparison correction tool to prevent false positives. Failing this gate means that when controlling for data mining bias, the model's performance is not statistically differentiable from random noise on almost all individual stocks.

\section{Results}\label{sec3}

Based on specific diagnostic metrics, our QRC models (both quantum and quantum-inspired) show forecasting potential beating AR baseline in all experiments and showing better results than GARCH baseline on 8/16 symbols: AAPL, IWM, MSFT, NVDA, QQQ, SPY, XLP, XLK.

In particular, our model wins on the biggest individual equities, but the 95\% CI for $\Delta$QLIKE crosses zero:

\begin{itemize}
    \item AAPL: 95\% CI (-0.225 to 0.0192) - Not significant.
    \item MSFT: 95\% CI (-0.203 to 0.043) - Not significant.
    \item NVDA: 95\% CI (-0.221 to 0.115) - Not significant.
    \item QQQ: 95\% CI (-0.270 to 0.001) - Not significant, boundary.
\end{itemize}

Because the upper bound of these intervals is positive, we cannot reject the null hypothesis that GARCH is just as good/better on these high-volume assets. 

\begin{figure}[h]
\centering 
\includegraphics[width=1.0\textwidth]{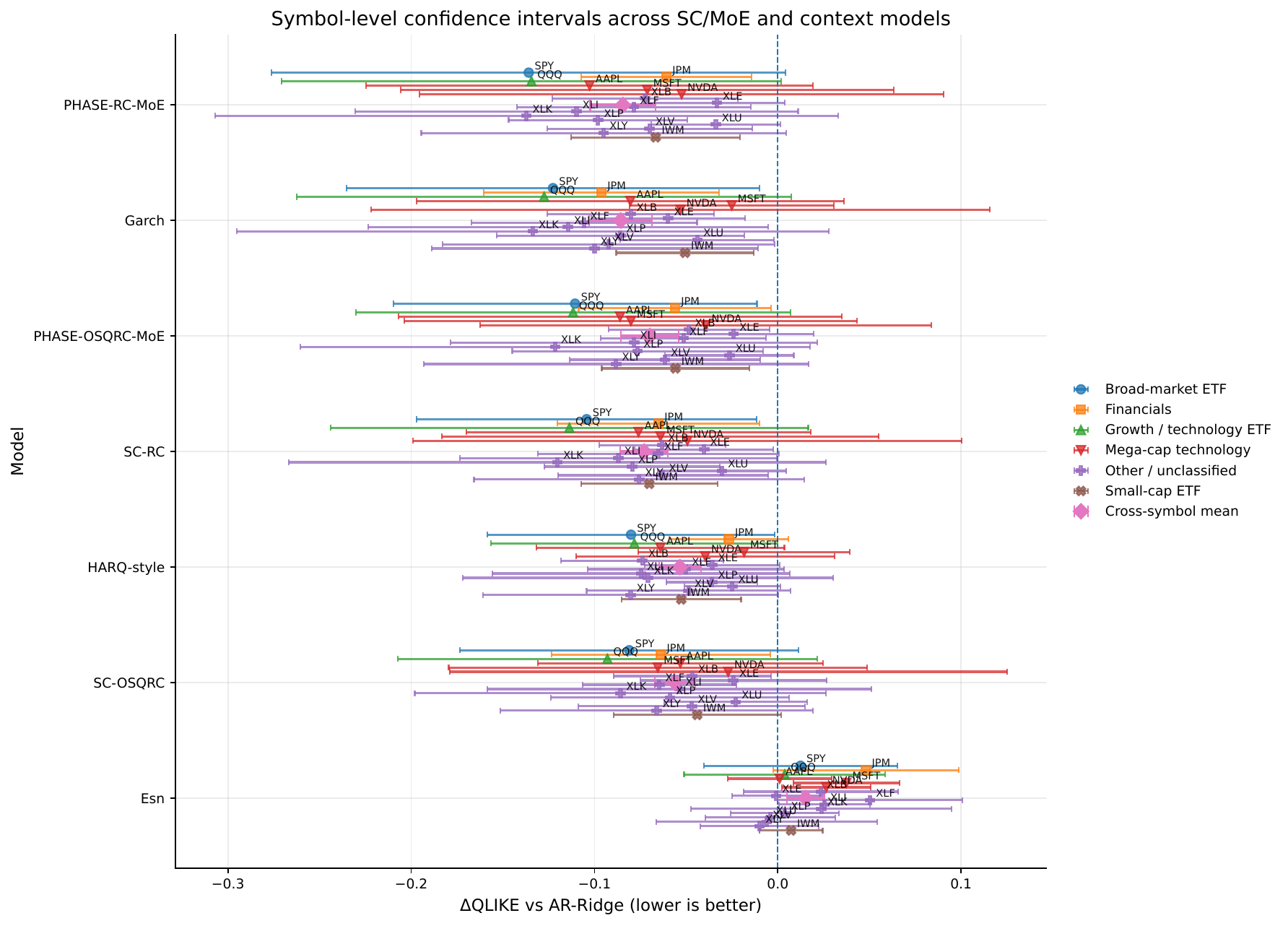}
\caption{Symbol-level confidence intervals for the QLIKE loss differential
relative to AR-Ridge. Each point denotes the estimated mean
$\Delta\mathrm{QLIKE}
=\mathrm{QLIKE}_{\mathrm{model}}-\mathrm{QLIKE}_{\mathrm{AR}}$
for one symbol, aggregated across the held-out folds, and each horizontal
bar shows its 95\% CI. Intervals crossing the vertical
zero line do not provide statistically detectable evidence of a difference.
Colors label the corresponding symbol class. The substantial
variation across individual equities and ETFs illustrates the cross-sectional
heterogeneity of forecasting performance and motivates evaluation across
multiple assets rather than reliance on a single aggregate result.}\label{fig:symbol_by_model}
\end{figure}

When averaged across different markets, PHASE-RC-MoE effectively did not show a statistically detectable average difference with GARCH (Figure~\ref{fig:symbol_by_model}). 

Rigorous statistical tests show that model performance was heterogeneous across markets, and detailed analysis revealed interesting behavior.

Models achieved a "Robust, consistent, tail-safe" evidence tier with a 95\% CI entirely below zero on two symbols, also significantly outperforming GARCH at the same time on all 3 folds over all held-out test periods:

\begin{itemize}
    \item IWM (Small Caps): $\Delta$QLIKE -0.070 (95\% CI -0.107 to -0.033)
    \item XLP (Consumer Goods): $\Delta$QLIKE -0.098 (95\% CI -0.146 to -0.049)
\end{itemize}

\begin{figure}[h]
\centering 
\includegraphics[width=1.0\textwidth]{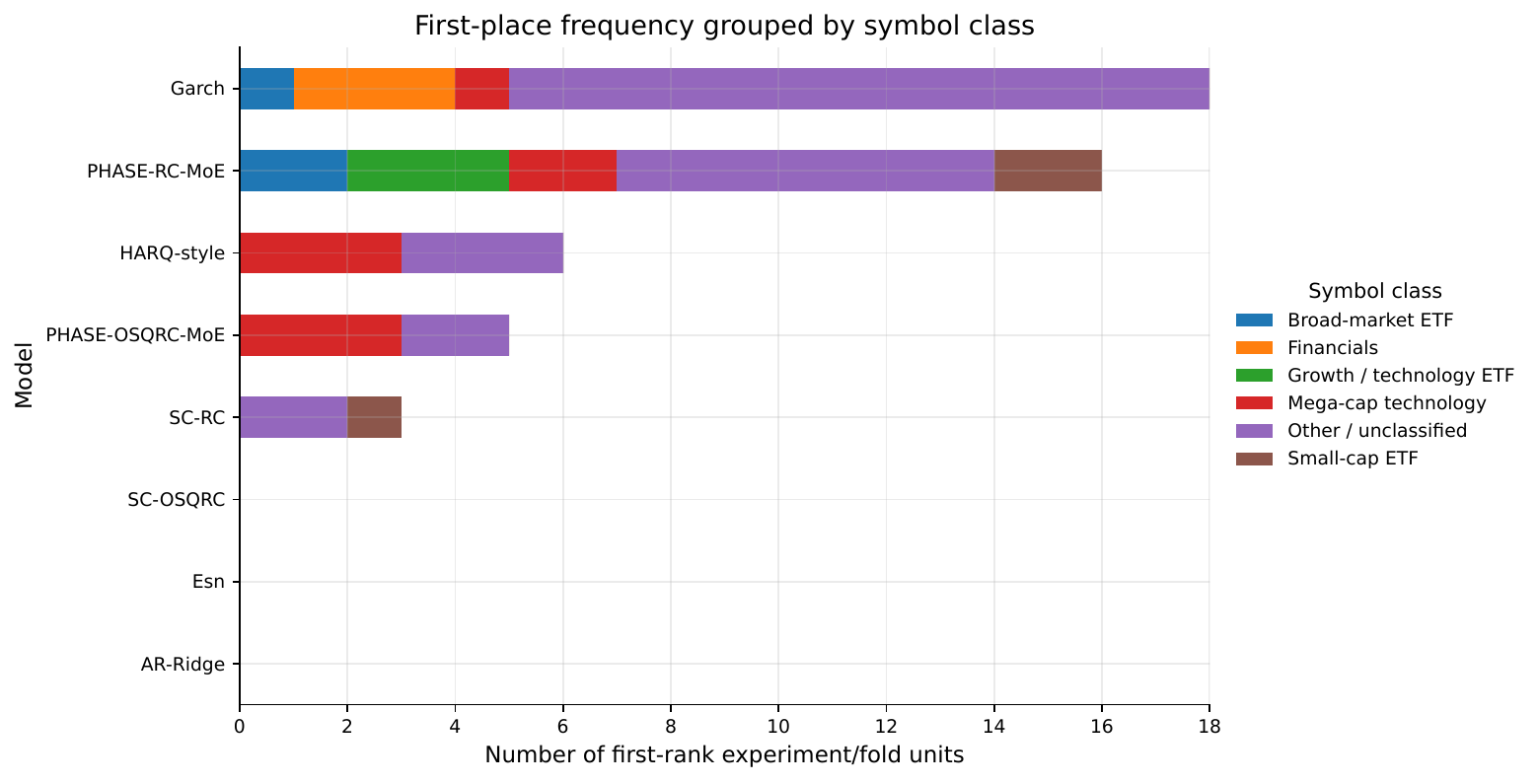}
\caption{Each bar represents the number of times when a model achieved the best QLIKE. Tied winners are counted as first-place occurrences.}\label{fig:first_place_by_model}
\end{figure}

These results provide evidence that the proposed architecture can outperform GARCH in specific market segments rather than uniformly across all markets (Figure~\ref{fig:first_place_by_model}). For robust claims we will need more tests across larger set of symbols and more targeted ablation tests.

To evaluate more accurately the potential for immediate usage of models to improve the accuracy of predictions via stacking, we calculated pointwise disagreement between SUSA models and GARCH/HARQ baselines. The results are summarized in Table~\ref{tab:stacking_results}.

\begin{table}[htbp]
    \centering
    \begin{tabularx}{\textwidth}{X c c c c}
        \toprule
        \textbf{Method} & \textbf{Stack QLIKE} & \textbf{Mean Gain} & \textbf{Win Share} & \textbf{DM p-value} \\
        \midrule
        SC-RC + HARQ & -7.762 & 0.0116 & 75.0\% & $< 0.001$ \\
        HARQ + PHASE-OSQRC-MoE & -7.763 & 0.0114 & 79.2\% & $< 0.001$ \\
        HARQ + PHASE-RC-MoE & -7.767 & 0.0077 & 75.0\% & 0.0046 \\
        \bottomrule
    \end{tabularx}
    \caption{Stacking performance compared to the best individual model member. Mean gain and win share highlight robust improvements, supported by Diebold-Mariano significance.}
    \label{tab:stacking_results}
\end{table}
  
The analysis reveals a forecast complementarity between HARQ and the proposed QRC architectures. The strongest deployable result is an equal-weight SC-RC–HARQ ensemble, which improves QLIKE over its better constituent by 0.0116 on average, wins in 75\% of symbol/fold units, and has a block-bootstrap confidence interval that entirely favors the stack. The equal-weight ensemble captures approximately 70\% of the improvement available to a test-fitted convex oracle, suggesting that most of the attainable diversification benefit can be obtained without unstable weight estimation.

\section{Conclusion}\label{sec4}

We introduced SUSA as a reservoir-design principle based on controlled response contrasts and regime-conditioned interpretation. We showed that these architectures can extract residual information that is not fully represented by a persistence anchor (such as AR) or by volatility-specific linear features. Performance is heterogeneous across markets: aggregate differences from GARCH are often statistically inconclusive, while the reported confidence intervals for IWM and XLP favor the proposed models. The complementarity observed with HARQ-style forecasts proves that the topological/dynamical features extracted by the complex-valued and periodic reservoirs, and/or quantum systems, are partially non-redundant and that the susceptibility representation captures information complementary to standard volatility features.

As a result, we can conclude with reasonable level of confidence that our empirical results validate the core hypothesis: these novel architectures can successfully isolate and extract residual information that linear volatility models miss.

SUSA models are stable and flexible: a susceptibility branch can be attached to a reliable classical anchor as a bounded correction, contributing value through forecast combinations even when its standalone advantage is modest. While our quantum implementations successfully extend this principle to open-system qubit dynamics and richer observable families, the present experiments do not yet establish a quantum advantage.

The mathematical richness of this architecture is now established; the primary challenge moving forward is identifying the optimal concrete implementations. Both the classical regime-conditioned reservoirs and their open-system quantum counterparts demonstrate immense theoretical potential for handling low signal-to-noise financial data. However, translating this potential into consistent dominance requires further refinement of model design. Future work must focus on studies to isolate the exact drivers of performance. The next question is no longer whether susceptibility architectures can model financial noise, but how to parameterize and physically implement them to fully unleash their advantage.

\section{Code availability}\label{sec5}

Code, split metadata, experiment configurations, random seeds, and scripts
used to reproduce the tables and figures will be released at
\url{https://github.com/akaliutau/qrc-susa-qbraid}. The repo also contains the detailed analysis of experiments, along with results themselves. 


\begin{appendices}

\section{Forecasting and Estimation Details}
\label{app:estimation}

Each outer fold contains chronological training, validation, and test
segments. Adjacent segments are separated by a purge of $H-1$ observations.
All scalers, stress thresholds, reservoir configurations, projections, and
readout parameters are estimated without access to the outer test segment.

The AR-Ridge anchor is
\begin{equation}
    \widehat y^{\,\mathrm{AR}}_i
    =
    \widehat\beta_0
    +
    \widehat{\boldsymbol\beta}^{\top}
    \operatorname{Std}_{\mathcal T}(\mathbf x_i).
\end{equation}
To avoid fitting the nonlinear correction to in-sample anchor residuals, its
training targets use expanding-window out-of-fold anchor predictions
$\widehat y^{\,\mathrm{AR,OOF}}_i$.

Reservoir features are standardized and projected to a validation-selected
rank before the direct-QLIKE readout in
Equation~\eqref{eq:qlike_readout}. The final correction is capped at
$C=1$, so the nonlinear component can modify but not replace the common
persistence forecast.

Candidate configurations are selected using
\begin{equation}
    S_{\mathrm{val}}
    =
    Q_{\mathrm{val}}
    +
    0.20
    \left|
        Q_{\mathrm{early}}-Q_{\mathrm{late}}
    \right|
    +
    0.10\max(\Delta Q_{\mathrm{calm}},0)
    +
    2\cdot10^{-5}d,
    \label{eq:validation_score}
\end{equation}
where $d$ is the retained feature dimension. The additional terms penalize
temporally unstable validation performance, deterioration in calm periods,
and unnecessarily large readouts.

\section{Classical Susceptibility Features}
\label{app:classical_features}

For a standardized scalar input $\widetilde x_t$, the fixed complex encoding is
\begin{equation}
    \mathbf d(\widetilde x_t)
    =
    \exp\!\left(
        \mathrm i\widetilde x_t\boldsymbol\phi
    \right)
    \odot
    \tanh\!\left(
        \widetilde x_t\boldsymbol\eta
    \right).
    \label{eq:complex_encoding}
\end{equation}
The open and ring reservoirs then follow
Equation~\eqref{eq:classical_reservoir}. Their transition matrices use
directionally asymmetric nearest-neighbour couplings,
\begin{equation}
    (\mathbf W_b)_{j+1,j}
    =
    \kappa(1+\chi)e^{\mathrm i\psi_j^+},
    \qquad
    (\mathbf W_b)_{j,j+1}
    =
    \kappa(1-\chi)e^{\mathrm i\psi_j^-}.
    \label{eq:directional_coupling}
\end{equation}
The ring contains the corresponding pair of wrap couplings between nodes
$q$ and $1$. Both matrices are rescaled to the same spectral radius.

For each topology, the observable vector contains final node populations,
nearest-neighbour currents, boundary imbalance, concentration, entropy, and
two temporal amplitude summaries:
\begin{equation}
\begin{split}
    \mathbf o^{(b)}
    =
    [&\{|s_j|^2\}_{j=1}^{q};
      \{\operatorname{Im}(\overline s_j s_{j+1})\}_{j=1}^{q-1};\\
     &|s_1|^2,\ |s_q|^2,\ |s_q|^2-|s_1|^2;\
      \operatorname{IPR};\ S;\
      \mu_t(|s_t|);\ \sigma_t(|s_t|)].
    \label{eq:classical_observables}
\end{split}
\end{equation}
The final model input is the signed and normalized open--ring contrast in
Equation~\eqref{eq:sc_representation}. When multiple fixed reservoir seeds
are used, their mean contrast and the cross-seed standard deviation are
concatenated.

The evaluated classical reservoir candidates are summarized below.
\begin{equation}
    (q,D,\chi,A)
    \in
    \{
      (5,2,0.25,1),
      (5,3,0.40,3),
      (10,2,0.35,1),
      (15,2,0.45,1)
    \},
    \label{eq:classical_grid}
\end{equation}
where $q$ is the number of nodes, $D$ the internal update depth, $\chi$ the
directional asymmetry, and $A$ the number of fixed seeds.

\subsection{Phase specialization}

The causal gate uses trailing origin-time statistics,
\begin{equation}
    \mathbf g_t
    =
    [
      h_d,h_w,h_m,
      \log RQ_5,\log RQ_{21},
      S^{RQ},J,DD_{63},
      I,T,K,
      VoV_5,VoV_{21},R
    ]^{\top}.
    \label{eq:gate_features}
\end{equation}
Here $I=h_d-h_m$, $T=h_w-h_m$, and
$K=h_d-2h_w+h_m$ encode the position, slope, and curvature of the
daily--weekly--monthly volatility state.

Each expert is fitted primarily on its own phase while retaining a small
background weight:
\begin{equation}
    w_i^{(m)}
    =
    \begin{cases}
      1, & m_i=m,\\
      0.15, & m_i\neq m.
    \end{cases}
    \label{eq:phase_weights}
\end{equation}
This gives phase specialization without making rare-phase readouts
ill-conditioned.

\section{Open-System Quantum Susceptibility Features}
\label{app:quantum_features}

The quantum input encoding is
\begin{equation}
    U_{\mathrm{in}}(x_t)
    =
    \bigotimes_{j=1}^{q}
    R_Z\!\left(
        \eta_{\phi}b_j x_t
    \right)
    R_Y\!\left(
        \eta_{y}a_j\tanh x_t
    \right),
    \label{eq:quantum_encoding}
\end{equation}
with fixed masks $a_j$ and $b_j$.

For edge set $E_b$, the coherent layer is
\begin{equation}
    U_{E_b}
    =
    \left[
      \prod_{(j,k)\in E_b}
      R_{YY}^{(j,k)}(\vartheta)
      R_{XX}^{(j,k)}(\vartheta)
    \right]
    \left[
      \prod_{j=1}^{q}R_Z^{(j)}(h_j)
    \right].
    \label{eq:quantum_coherent}
\end{equation}
This layer is followed by directional excitation transfer with probabilities
$p_{\rightarrow}>p_{\leftarrow}$ and by local amplitude damping with rate
$\gamma$, producing the complete map in
Equation~\eqref{eq:quantum_reservoir}.

The measured quantities are local $Z$ expectations and adjacent
$XX$, $YY$, $XY$, and $YX$ correlations. They are converted into
\begin{align}
    n_j
    &=
    \frac{1-\langle Z_j\rangle}{2},\\
    C_j
    &=
    \frac{
        \langle X_jX_{j+1}\rangle
        +
        \langle Y_jY_{j+1}\rangle
    }{2},\\
    J_j
    &=
    \frac{
        \langle X_jY_{j+1}\rangle
        -
        \langle Y_jX_{j+1}\rangle
    }{2}.
    \label{eq:quantum_observables}
\end{align}
The topology-specific vector contains the final, temporal-mean, and
temporal-standard-deviation summaries of populations, coherences, currents,
boundary imbalance, concentration, and entropy:
\begin{equation}
    \mathbf z^{(b)}
    =
    [
      \mathbf o_L^{(b)};
      \operatorname{mean}_{t}\mathbf o_t^{(b)};
      \operatorname{sd}_{t}\mathbf o_t^{(b)}
    ].
    \label{eq:quantum_summary}
\end{equation}

SC-OSQRC uses
\begin{equation}
    \mathbf z^{\mathrm{SC-Q}}
    =
    \left[
      \mathbf z^{(o)}-\mathbf z^{(r)};
      \frac{
        \mathbf z^{(o)}-\mathbf z^{(r)}
      }{
        |\mathbf z^{(o)}|+|\mathbf z^{(r)}|+\varepsilon_z
      }
    \right].
    \label{eq:quantum_sc_features}
\end{equation}
The phase-conditioned quantum model additionally retains the two absolute
topology summaries:
\begin{equation}
    \mathbf z^{\mathrm{PHASE-Q}}
    =
    [
      \mathbf z^{(o)};
      \mathbf z^{(r)};
      \mathbf z^{\mathrm{SC-Q}}
    ].
    \label{eq:quantum_phase_features}
\end{equation}

The evaluated quantum dynamics are
\begin{equation}
\begin{split}
(q,\eta_y,\eta_{\phi},\vartheta,h,
 p_{\rightarrow},p_{\leftarrow},\gamma)
\in\{&
(3,0.55,0.25,0.24,0.12,0.10,0.020,0.010),\\
&
(4,0.42,0.22,0.18,0.10,0.08,0.015,0.008)
\}.
\end{split}
\label{eq:quantum_grid}
\end{equation}

All reported quantum features are obtained by exact density-matrix simulation.
The quantum dynamics are fixed; model selection and all trainable readouts are
classical.

\end{appendices}

\bibliography{lit-li}

\end{document}